\documentclass[runningheads]{llncs}
\usepackage[T1]{fontenc}
\usepackage[position=top]{subfig}
\usepackage{floatrow}  
\usepackage{caption}
\usepackage{dingbat}
\usepackage{amsfonts}

\usepackage{amsmath,graphicx}
\usepackage[position=top]{subfig}
\usepackage{xcolor}

\usepackage{soulutf8}
\sethlcolor{yellow!50}

\newcommand{\highlightx}[1]{#1}
\begin{document}
\title{Learning Object Focused Attention}
\author{Vivek Trivedy\inst{1} \and
Amani Almalki\inst{1} \and
Longin Jan Latecki\inst{1}}
\authorrunning{V. Trivedy et al.}
\institute{Temple University, Philadelphia, PA 19122, USA \\
\email{\{vivek.trivedy,amani.almalki,latecki\}@temple.edu}}
\maketitle              % typeset the header of the contribution
\begin{abstract}
We propose an adaptation to the training of Vision Transformers (ViTs) that allows for an explicit modeling of objects during the attention computation. 
This is achieved by adding a new branch to selected attention layers that computes an auxiliary loss which we call the object-focused attention (OFA) loss.
We restrict the attention to image patches that belong to the same object class, which allows ViTs to gain a better understanding of configural (or holistic) object shapes by focusing on intra-object patches instead of other patches such as those in the background. 
Our proposed inductive bias fits easily into the attention framework of transformers since it only adds an auxiliary loss over selected attention layers. Furthermore, our approach has no additional overhead during inference.
We also experiment with multiscale masking to further improve the performance of our OFA model
and give a path forward for self-supervised learning with our method.
Our experimental results demonstrate that ViTs with OFA achieve better classification results than their base models, exhibit a stronger generalization ability to out-of-distribution (OOD) and adversarially corrupted images, and learn representations based on object shapes rather than spurious correlations via general textures.  For our  OOD setting, we generate a novel dataset using the COCO dataset and Stable Diffusion inpainting which we plan to share with the community.

\keywords{representation learning  \and vision transformers \and attention mechanism.}
\end{abstract}

\section{Introduction}
\label{sec:Intro}
One of the key ideas of vision transformers (ViTs) is to update the representation of a given patch $p$ as a weighted sum of feature vectors from all image patches. The weights, which are computed using the transformer attention mechanism, are determined based on the feature similarity of $p$ to other patches. 
This is based on an implicit assumption that features of patches within the same object should be more similar to each other than to features of other objects or of the background. However, this assumption is often not satisfied, since different parts of the same object may have very different appearances, and some object patches may be more similar to the background or other object patches. This fact limits learning efficiency and also the generalization ability on both in distribution and out of distribution samples.  In addition, ViTs are susceptible to learning "shortcuts" \cite{shortcuts} where rather than capturing the object focused semantic meaning of an image, they capture spurious correlations with the background or other image artifacts.
For example, if all training images show a fox in the forest, then a fox on a street may not be recognized. Currently, this problem is alleviated with a large number of training images via datasets such as ImageNet and heavy data augmentation. The hope is that the fox will appear on a large variety of backgrounds, but the assurance of this fact comes only from a large number of images, and
it is hard to guess what other anomalies may be hidden in the training images. 

Our key contribution is to limit the attention of patches to patches of the same object class only in a learned way. 
It can be viewed as refocusing the attention on relevant image parts. As demonstrated in \cite{FocalModNet2022}, such an approach can lead to significant performance improvement.  
However, the focal modulation in \cite{FocalModNet2022} is done outside the transformer framework, and it does not include any inductive bias to focus on patches of the same object class, as proposed here. 

\begin{figure}
\centering
\includegraphics[width=0.7\linewidth]{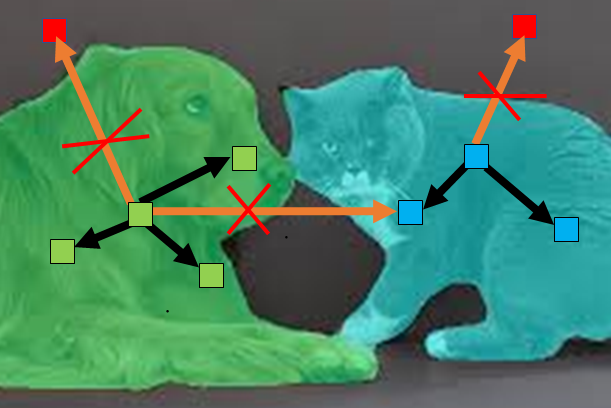}
\caption{We restrict learning attention to objects of the same class.}
\label{fig:catseg}
\end{figure}

We illustrate our idea on an example image in Fig.~\ref{fig:catseg}. We propose to limit the attention of the green patch $p$ inside the dog to patches inside the green mask. Hence, the red patch in the background and the blue patch inside the cat in the blue mask are excluded from computing the new weighted representation of the green patch.
%Similarly, the attention of the red patch on the grass is limited to other red patches on the grass.
Furthermore, our proposed restriction on the patch attention is not hard coded but learned. This is achieved by adding a new branch to selected attention layers that computes an auxiliary loss called object focused attention (OFA) loss.  To train ViTs with the proposed semantically focused attention, we use datasets with semantic segmentation masks. Luckily there exists a plethora of such datasets like the MS COCO dataset or PASCAL VOC 2012 dataset. 
In the absence of segmentation masks out of the box, we note the ability to generate psuedo-segmentation masks via general purpose segmentation models such as the Segment Anything Model (SAM) \cite{segmentanything}.

The proposed restriction of attention to patches within the same object allows transformers to gain a better understanding of configural (or holistic) object shapes since attention is trained to be learned within patches of the same object class, hence the background is largely ignored. 
This also means better generalization to out-of-distribution (OOD) images.
% and higher resistance to adversarial attacks.
We present an experimental evaluation to demonstrate these facts
on multilabel classification tasks.
%For this, we will evaluate the performance of the proposed approach on various downstream tasks from image classification through object detection to instance segmentation. 
As our baseline model, we use the Musiq Transformer \cite{ke2021musiq} and also show results with strong out-of-distribution performance with the standard ViT \cite{ViT}.  We chose Musiq Transformer due to its 2D positional encoding that is suitable for multiscale image representation. The original Musiq Transformer is developed for image quality assessment, but we adopt it for other downstream tasks such as multilabel classification.
We note that our proposed OFA branch can be easily added to the self-attention layer of any vision transformer variant.

\section{Object Focused Attention}
%\subsubsection{T1-1: Restricting the transformer attention to objects and stuff}
As outlined in Section \ref{sec:Intro}, our key idea is to limit the attention of patches to patches of the same class. Here we introduce our formal framework to implement this idea.

For ViT and its variants, an input image $I$ is first divided into $N$ disjoint square patches $\mathcal{P}=\{p_1, \dots, p_N\}$ of a fixed size.
For simplicity of presentation,
we focus on a single encoder layer of Musiq \cite{ke2021musiq} with one head.
Let $\{ \mathbf{x}_1, \dots, \mathbf{x}_N \}$ be the set of input tokens representing the patches that were obtained by the previous layer, where each token is a row feature vector $\mathbf{x}_i \in \mathbb{R}^d$.
Let $\mathbf{X} \in \mathbb{R}^{N \times d}$ be the matrix obtained by stacking vectors $\mathbf{x}_1 \dots \mathbf{x}_N$.
The scaled attention module of this layer
first linearly projects the patch tokens to query, key, and value matrices $\mathbf{Q,K,V} \in \mathbb{R}^{d \times N}$, given by
$\mathbf{Q}=\mathbf{X}\mathbf{W}_Q$,
$\mathbf{K}=\mathbf{X}\mathbf{W}_K$,
$\mathbf{V}=\mathbf{X}\mathbf{W}_V$,
where $\mathbf{W}_Q, \mathbf{W}_K, \mathbf{W}_V \in \mathbb{R}^{d \times d}$
are learnable parameter matrices.

Next, we compute the attention weight matrix $\mathbf{A}$ that reflects the similarity between the patches:
\begin{equation}
    \mathbf{S} = \frac{\mathbf{Q K}^T}{\sqrt{d}} 
    \ \ \ \ \ \text{and} \ \ \ \ \
    \mathbf{A} = softmax(\mathbf{S}) \in 
    \mathbb{R}^{N \times N}.
\end{equation}
We call matrix $\mathbf{S}$ a scaled pre-attention matrix. 
The $i$-th row of $\mathbf{S}$ is denoted as $\mathbf{s}_i \in \mathbb{R}^{1 \times N}$, and it indicates the attention of patch $i$ to all other patches.

Finally, the output matrix is obtained as
$\mathbf{Y} = \mathbf{A V} \in \mathbb{R}^{N \times d}$,
where each row $\mathbf{y}_i$ of matrix $\mathbf{Y}$ is a new representation  of patch $\mathbf{x}_i$ as the sum of vectors in $\mathbf{V}$ weighted by $i$-th row $\mathbf{a}_i$ of attention matrix $\mathbf{A}$.  The new representation of the $i$-th patch token
is a weighted sum of all patch tokens.

\begin{figure}
\centering
%(a) \includegraphics[width=0.41\textwidth]{AttenV3.png} \hspace{20mm} (b) 
\includegraphics[width=0.5\textwidth]{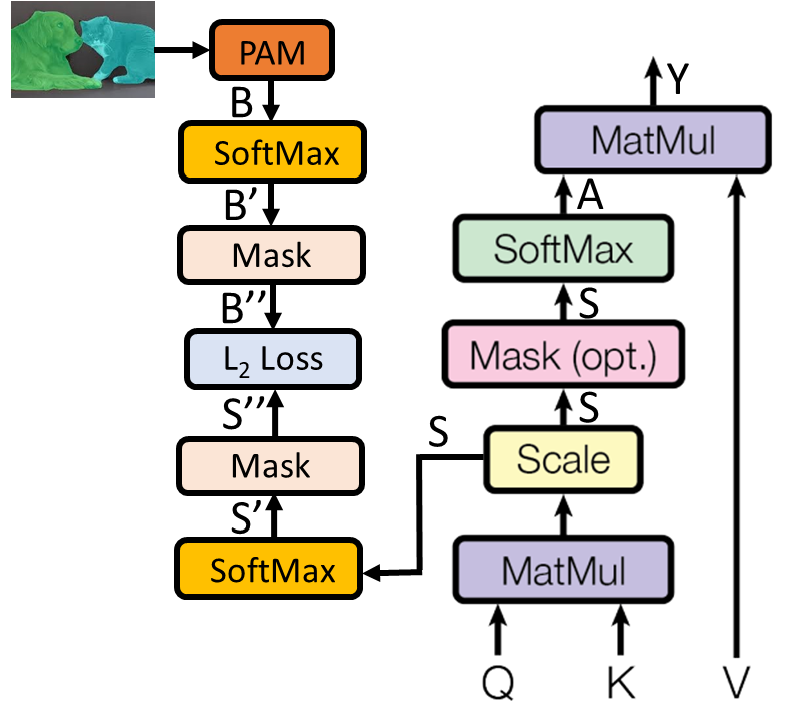}
\caption{The proposed object focused attention (OFA) as an extension of self-attention. The arrows are labeled with the input/output matrices. The right part of the diagram is based on the original self-attention paper \cite{AttentionNIPS2017}.
The left branch computes the OFA loss.
The patch adjacency matrix (PAM) module is used to compute the patch adjacency matrix $\mathbf{B}$, which is then compared to the pre-attention matrix $\mathbf{S}$.   
}
\label{fig:Aten}
\end{figure}

The right branch of the diagram in Fig.~\ref{fig:Aten} illustrates this process,
which is the standard attention computation as proposed in \cite{AttentionNIPS2017}.
The left branch of the diagram in Fig.~\ref{fig:Aten}
illustrates the proposed object focused attention (OFA) 
that aims at training matrix $\mathbf{S}$ to resemble a binary matrix $\mathbf{B}$ representing a focus on patches within a given object. 
The left branch is devoted to computing OFA loss.
The matrix $\mathbf{B}$ and the process of computing OFA loss are defined below.

\begin{figure*}[h]
\centering
\includegraphics[scale=0.5]{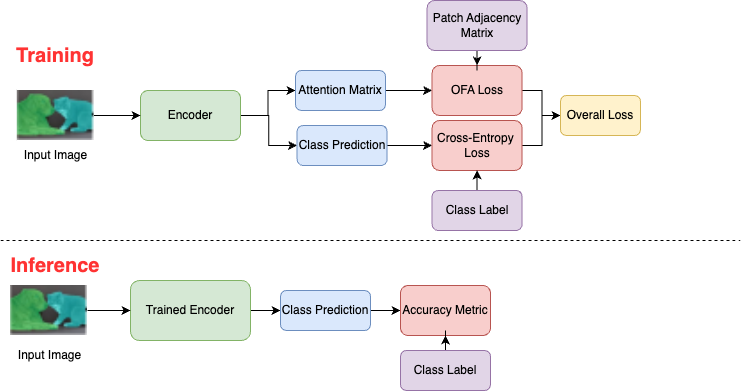}
\caption{\highlightx{Data flow showing differences in training and inference. OFA is shown explicitly as a training time method and thus can be used without any segmentation labels during inference.}
}
\label{fig:dataflow}
\end{figure*}

Let $\mathcal{R}=\{R_1, \dots, R_r\}$ be a semantic segmentation of image $I$ into a set of disjoint regions (object masks) such that their union covers the whole image. We also assume that patch $p_i$ is contained in or intersects region $R_j$.
Our training procedure seeks to reduce the attention values of patches disjoint with region $R_j$ to zero in the pre-attention vector $\mathbf{s}_i$. We note that simply setting these values to zero for the training image $I$ will not generalize to test images for which no segmentation masks are given.  Therefore, we propose to learn this behavior by incorporating an auxiliary loss function to focus the attention of patch $i$ only on patches that also intersect region $R_j$. 
For this, we define a patch attention matrix (PAM) $\mathbf{B}$, which is a binary $N \times N$ matrix.
Ones in row $\mathbf{b}_i$ of $\mathbf{B}$ represent patches that intersect the same object mask as patch $i$. Formally,
$\mathbf{b}_{ik}=1$ if both patches $p_i$ and $p_k$ intersect the same object mask and zero otherwise.  We use here a simplified notation for clarity of presentation. In particular, patch $p_i$ may intersect more than one object mask $R_j$, in which case more regions need to be considered.  \highlightx{To handle overlap patches, we use a simple heuristic where if any part of a patch is part of an object, it is considered an object patch.}

Then we apply row-wise softmax to $\mathbf{B}$ and obtain $\mathbf{B'}=sofmax(\mathbf{B})$.
Since we want the patch cross attention to focus on foreground objects, we mask all rows in $\mathbf{B'}$ that represent background patches. We denote the new matrix $\mathbf{B''}$.

Similarly, we compute row-wise softmax to obtain $\mathbf{S'}=sofmax(\mathbf{S})$.
Followed by setting to zero (masking) all rows in $\mathbf{S'}$ that represent the background patches. The resulting matrix is denoted with $\mathbf{S''}$. 
We use matrices $\mathbf{S''}$ and $\mathbf{B''}$ to define the object focused attention (OFA) loss as their $L_2$ distance:
\begin{equation}
\label{eq:objectattloss}
\mathcal{L}_{OFA} = ||\mathbf{S''} - \mathbf{B''}||_2.
\end{equation}
This process is graphically illustrated in the left part of the diagram in Fig.~\ref{fig:Aten}.
We call the transformer trained with this auxiliary OFA loss \textbf{OFAMusiq}.

In order to explain the intuition behind OFA loss, let us assume that
row $i$ of $\mathbf{B}$ represents an object patch and has $k$ ones,
meaning there are $k$ other patches that intersect the same region.
Then $sofmax(\mathbf{B})$ maps the ones in row $i$ of $\mathbf{B}$ to $1/k$ in $\mathbf{B'}$,
and the same values will remain in $\mathbf{B''}$.
Hence the $L_2$ distance between rows $i$ of $\mathbf{B''}$ and $\mathbf{S''}$ pushes patch $i$ to pay equal attention to the other $k$ patches of the same object and zero attention to all other patches.
With reference to Fig.~\ref{fig:catseg}, 
OFA loss forces the green patch inside the dog to pay attention only to patches inside the green dog region.

Moreover, since the sum of each row of $\mathbf{B''}$ is one, the contribution of each patch to OFA loss is equal. This means that 
a patch $i$ that belongs to a small object, and hence has fewer neighbors in its attention graph (fewer ones in $i$ row of $\mathbf{B}$) is equally important as patches that belong to large objects.  

The proposed OFA loss can be placed at any layer or at several layers at the same time.
In Section~\ref{sec:experiments}, we explore options for the best placement of the OFA loss.
%Should it be applied only after the last layer, or at the first layer, or at the intermediate layers?
Our overall loss function can be summarized as follows:
\begin{equation}
\label{eq:overalllossViTVffffV}
\mathcal{L}_{total} = \mathcal{L}_{task} + \alpha \cdot \mathcal{L}_{OFA},
\end{equation}
where $\mathcal{L}_{task}$ is a task-dependent loss,
e.g., cross-entropy for classification,
and $\alpha$ is a hyperparameter that balances the two loss functions.

\section{Self-Supervised Option with MAE}
Our method uses datasets with semantic segmentation masks to train vision transformers with the proposed semantically focused attention.
While there exist many such datasets such as the MS COCO dataset, PASCAL VOC 2012, or PACO \cite{ramanathan2023paco},
they are relatively small, so we explore the setting with self-supervision which is useful in learning representations for low-data domains.  For this, we show experiments where we integrate OFA and Musiq Transformer with Masked AutoEncoder (MAE)\cite{MAE_CVPR2022}.

MAE uses self-supervised learning masking, where certain patches of an image are masked, and the model is tasked with predicting the original content within those masked regions. This approach encourages the model to learn meaningful representations by leveraging contextual information from the surrounding visual context.  The advantages of MAE lie in its ability to capture rich contextual dependencies and learn robust visual representations. Training the model to predict masked regions forces the model to understand and utilize the relationships and patterns present in the small amount of labeled data.

\begin{figure}
\centering
\includegraphics[width=0.7\textwidth]{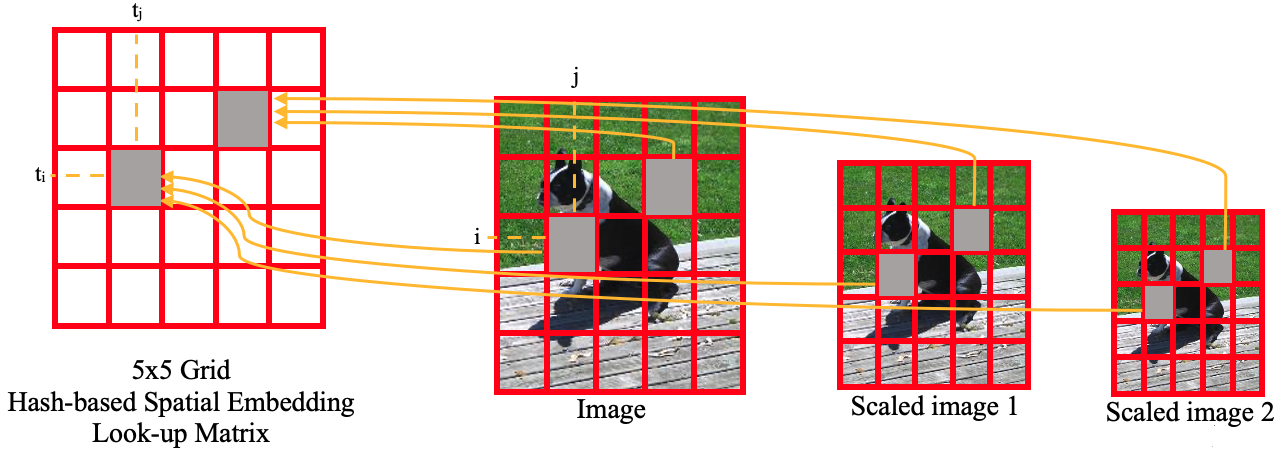}
\caption{The multiscale masking is computed by masking the grid cells (left) and carrying over the masked cells to image patches that correspond to those cells.}
\label{fig:MusiqMAE}
\end{figure}

%how is MAE integrated into Musiq? 
To our knowledge, we are the first ones to extend MAE to multiscale masking by utilizing Musiq positional encoding. 
Instead of performing masking directly on image patches, we propose to perform masking on the cells of the reference grid, which is then carried to tokens of images of different scales using a simple geometric mapping of the cell grids to image patches, see Fig.~\ref{fig:MusiqMAE}.  This mapping is used by Musiq for positional encoding, but we extend it to also guide the masking process.

\section{Adjacency Regularization} 
\label{sec:adjacencyreg}
Another way to view our OFA loss is through the lens of "adjacency regularization" by enforcing a penalty on allowed states of connectivity.  Vanilla transformers such as ViT are known for their $O(N^2)$ quadratic complexity with respect to attention computation over the number of input patches.  Cast in the language of graphs, this is a complete graph (with self-loops) where every pair of vertices is connected via an edge producing $\frac{N\cdot(N+1)}{2}$ edges of order $O(N^2)$.  The adjacency matrix, $\mathbf{B}$, for such a graph can be described as $\mathbf{B}_{ij}=1 \ \forall i,j \in \{0, \dots ,N-1\}$, where $N$ is the number of vertices in the graph, or in the case of ViT, input patches.  
By restricting the attention of each object patch to only patches of the same object, we significantly reduce the number of edges in the attention graph represented by matrix $\mathbf{B}$.  In particular for MS COCO \cite{coco}, purely object based connectivity creates a roughly $80\%$ reduction in the number of edges.  Only $20.7\%$ of edges from the 
standard fully-connected attention are used. The underlying motivation behind training to decrease connectivity is to encourage a more parsimonious attention matrix which is robust to spurious correlations and instead can focus on semantic object information \cite{parsimonious}.  We show empirically in Section \ref{sec:experiments} that our model achieves such robustness.

\section{Related Work}
\subsection{Transformers and Self-Attention}
The transformer's \cite{AttentionNIPS2017} self-attention mechanism offers a way for allowing every token to model information over every other token.  ViT \cite{ViT} adapted the transformer for computer vision by converting an image to a set of patch tokens and then using the standard transformer blocks.

% While the self-attention-based vision transformers have outperformed CNNs on most benchmarks like ImageNet dataset \cite{ImagenetCVPR2009}, the computational cost of stacked
% attention blocks with multiple heads is large, and it grows quadratically with the number of patches.  Therefore, much research has focused on reducing the computational cost of self-attention in ViT. 
% \cite{TokenMergeICLR2022} proposes to improve efficiency by identifying attentive tokens and merging unattentive tokens and hence reducing the number of tokens used in the attention layers.
% In \cite{VaswaniCVPR2021}, the authors propose a new self-attention mechanism called Local Self-Attention (LSA) that reduces the computational cost of self-attention by using a fixed, pre-defined set of attention weights for each input patch. This approach reduces the number of attention parameters, making the model more efficient and easier to train.
% \cite{MengCVPR2022} introduces AdaViT, an adaptive computation framework that learns to derive usage policies for patches, self-attention heads, and transformer blocks to use throughout the backbone with the aim to improve inference efficiency. A lightweight decision network is attached to the backbone to produce decisions on-the-fly.
% DynamicViT \cite{DynamicvitNIPS2021} proposes a dynamic token sparsification to prune the tokens. This model stresses the importance of information interactions with heuristic patch-wise interactions.

There has been a considerable amount of work related to improving the self-attention mechanism and augmenting the inductive biases in vision transformers.
One line of research has focused on modifying the self-attention mechanism to better capture spatial information in images. 
For example, \cite{ViTattentionNIPS2021} suggests using a mixture of local and global tokens in the input embedding to improve the model's ability to capture both local and global information in the image.
Swin transformer \cite{liu2021swin} utilizes a hierarchical structure analog to Convolutional Neural Networks (CNNs) to improve ViT performance.
Learning of attention has been considered in \cite{JieMa2022}, where it is applied to rectangular windows of patches. Since the size of the windows is learned, the approach is called window-free multi-head attention.
%In contrast to our work, no object information has been considered in those approaches.
In contrast to our work, all these approaches do not explicitly utilize object mask knowledge in restricting or restructuring self-attention. Moreover, many of them add computational overhead at the time of inference while our approach keeps the original structure of the self-attention layer during the inference.

\subsection{Holistic Shape Representation}

According to \cite{Configural2022}, objects have both local and configural shape properties.
Local shape properties can be important for recognition.  For example, ears alone may be sufficient to identify a rabbit but often are not discriminative enough.
A configural shape property is a function not just of one or more local features (parts) but also of their arrangement meaning it provides a holistic shape representation.

Vision transformers, like other deep learning models, can learn to attend to different features of an image, including both texture and shape. However, it has been observed that their attention is more focused on texture than shape, in particular, they fail to capture the configural nature of shapes in images, which means they are not able to adequately learn a holistic shape representation \cite{Configural2022}.
There have also been several studies on CNNs that demonstrated that they tend to attend more to texture than shape in natural images, e.g., \cite{Kellman2018,GeirhosNeurIPS2018,texture-shape-ICLR2019}.
We demonstrate in Section~\ref{sec:experiments} that the proposed refocusing of attention within objects contributes to a better understanding of the holistic shape of objects.

\subsection{Multi-Label Classification}
In many classification tasks, class labels are mutually exclusive such as when an image contains just one object.  In multi-label classification, we predict
mutually non-exclusive class labels,
such as when an image may contain more than one object or concept.  Multilabel classification is a challenging problem in computer vision due to the high dimensionality of the label space and potential correlations between labels. The label space can contain a large number of labels, and each label can be associated with multiple instances in the dataset. Furthermore, the labels can be highly correlated, meaning that the presence of one label in an image can increase the likelihood of other labels being present as well.

One of the first transformer networks applied to multilabel classification is \cite{MlTr2022}, where
windows partitioning, in-window pixel attention, and cross-window attention are used for improving the performance of multi-label image classification tasks. 
One of the best-performing multilabel classification method is ADDS~\cite{MLClass2022}, where ADDS stands for 
Aligned Dual moDality ClaSsifier.
It includes a dual-modal decoder that performs alignment between visual and textual features.  In contrast, we only use visual features.

\section{Experimental Evaluation} 
\label{sec:experiments}

Across our experiments, we use both single-scale and multi-scale MUSIQ transformers \cite{ke2021musiq}, denoted 
MUSIQ-single and MUSIQ-multi.
The single-scale resizes images so that the longer side has length 512 while preserving the aspect ratio (ARP). The multi-scale uses the full-size image and two ARP resized inputs 384 and 224. It, therefore, uses three-scale input. In addition, we investigate the influence of self-supervised learning using MAE masking as a further enhancement of our methods.  We also show that OFA is much more robust to background perturbations than standard ViTs by evaluating on our Stable Diffusion inpainted dataset.  \highlightx{We use $\alpha=0.7$ across our experiments unless otherwise stated.}  Finally, we present an interesting finding via patch shuffling showing that ViTs don't grasp the overall shape of objects well compared to models equipped with OFA.

\subsection{Multi-Label Classification on MS-COCO and Pascal Voc2012} \label{sec:COCOclass}

MS COCO (Microsoft Common Objects in Context) is a large-scale image recognition dataset containing 80 different object categories.
Multilabel classification uses the same train/val splits as for the object detection task. The training set contains 118,287 images with annotations, while the validation set contains 5,000 images, which are used for testing. All the training images also contain semantic segmentation masks so that we can use them in our framework.  \highlightx{We use the standard definition given by the COCO dataset of \textit{thing} and \textit{stuff}. From the COCO homepage we quote: "Things are objects with a specific size and shape, that are often composed of parts. Stuff classes are background materials that are defined by homogeneous or repetitive patterns of
fine-scale properties, but have no specific or distinctive spatial extent or shape."
Put simply the COCO dataset defines segmentation masks directly for object
classes and background classes.}

Pascal VOC 2012 \cite{pascal-voc-2012} contains objects grouped into 20 classes.
The standard train/val set for the multilabel image classification/detection task has 11,540 images.
However, since we need semantic segmentation masks, we train on train/val 2,913 images that are usually used for the image segmentation task.
We test on the standard Pascal VOC 2012 test set composed of 10,991 images.  Following other methods, we use mean average precision (mAP) in evaluating multilabel classification performance.

We experiment with computing the OFA loss over multiple attention layers of MUSIQ, which has 14 attention layers.
Table \ref{tab:position} compares two settings for positioning the OFA loss: at the first and last layers [1, 14] and at layers $[1,7,14]$. 
Since placing OFA loss at layers $[1,7,14]$ performs the best across all the settings, this model is used in all our further experiments.
As for our weighting schema, we progressively weight the contributions of each attention block with later layers getting more weight with a factor of $0.9$. The loss at layers $[1,7,14]$ is weighted as: 
\begin{equation}
    OFA_{total} = \frac{1}{3}(0.9 \cdot OFA_{14} + 0.9^2 \cdot OFA_{7} + 0.9^3 \cdot OFA_{1})
\end{equation}
 The loss at layers $[1,14]$ is weighted as: 
\begin{equation}
    OFA_{total} = \frac{1}{2}(0.9 \cdot OFA_{14} + 0.9^2 \cdot OFA_{1})
\end{equation}

\begin{table}[]
\centering
\begin{tabular}{c|cccc}
Methods  & \multicolumn{2}{c}{MS COCO}  & \multicolumn{2}{c}{PASCAL VOC2012} \\ 
&  [1,14] & [1,7,14] & [1,14] & [1,7,14] \\ \hline
MUSIQ-single + OFA     & 88.3         & 89.0  & 87.8    &  88.4  \\
MUSIQ-multi + OFA     & 89.4      & 89.9  &   89.3   & 90.1  \\
MUSIQ-single + MAE + OFA    & 91.3        & 91.7  & 90.8    &  91.5 \\
MUSIQ-multi + MAE + OFA   & 91.6      & 92.1  &  91.2 & 91.9  \\
ViT-Base + OFA & 88.2& 89.0 & - & 87.8
\end{tabular}
\caption{mAP multi-label classification results for placement of the OFA across layers. Placing OFA loss at layers $[1,7,14]$ performs the best across all MUSIQ settings and so is used in further experiments. \highlightx{We add ViT and note that we use layer 12 instead of 14 as ViT-Base has 12 layers.}} 
\label{tab:position}
\end{table}

Table~\ref{tab:musiq} shows multilabel classification results of MUSIQ transformer trained on MS COCO.
We evaluate it on MS COCO and on Pascal VOC2012.
The results on Pascal VOC2012 can be interpreted as zero-shot since do not train the model on this dataset and instead just fine-tune a classification head. We only benefit from the fact that the 20 classes of Pascal VOC2012 are a subset of the 80 classes of MS COCO. However, these datasets are composed of disjoint images, and MS COCO images are very different from Pascal VOC2012 images. Hence the excellent performance of MUSIQ-multi + MAE + OFA gives an initial result showing out-of-distribution (OOD) generalization ability of our approach.

\begin{table}[t]
\centering
\begin{tabular}{c|cc}
Methods  & MS COCO  & zero-shot VOC2012 \\ \hline
\highlightx{ViT-Base} & 86.6 & 81.7\\
\highlightx{ViT-Base + OFA} & 87.3 & 87.8\\
MUSIQ-single   & 87.5        & 89.7 \\
MUSIQ-multi    & 88.0      & 90.2 \\
MUSIQ-single + OFA  & 89.0      & 90.9 \\
MUSIQ-single + MAE   & 89.7       & 92.3 \\
MUSIQ-multi  + OFA  & 89.9     & 93.2   \\
MUSIQ-multi  + MAE    & 91.6      & 93.6 \\
MUSIQ-single + MAE + OFA    & 91.7   & 94.7 \\
MUSIQ-multi  + MAE + OFA  & \textbf{92.1}   & \textbf{95.4} \\ 
\end{tabular}
\caption{mAP multilabel classification results on the MS COCO and Pascal VOC2012 datasets. All models are trained and evaluated on MS COCO. They are then applied on Pascal VOC2012 without any finetuning besides the linear head.} 
\label{tab:musiq}
\end{table}

In Table \ref{tab:sota}, we compare our methods to other multi-label classification methods on MS COCO, most with more complex architectures. We find that our method which adds an auxiliary loss to MUSIQ transformers outperforms other SOTA methods. We do not compare against multimodal methods such as \cite{MLClass2022} and \cite{mldecoder} since we only use visual features.  

\begin{table}[]
\centering
\begin{tabular}{c|cc}
Methods    & Resolution & mAP  \\ \hline
IDA-R101 \cite{liucausality} & 	576 & 86.3 \\
TResNet-XL \cite{ridnik2021asymmetric} & 640 & 88.4 \\
TResNet-L-V2 \cite{ridnik2021imagenet} &  640 & 89.8 \\
MlTr-XL \cite{MlTr2022}     & 384        & 90.0 \\
IDA-SwinL \cite{liucausality} & 384    & 90.3 \\
Q2L-SwinL \cite{query2label}    & 384       & 	90.5 \\
MLD-TResNet-L-AAM \cite{sovrasov2022combining}      & 640       & 91.3 \\
Q2L-CvT \cite{query2label} & 384       & 91.3 \\ \hline
\highlightx{MUSIQ-multi} & (full,384,224) & 88.0 \\
MUSIQ-multi  + MAE + OFA    &   (full,384,224)     & \textbf{92.1}  
\end{tabular}
\caption{Comparison to other methods on MS COCO. \highlightx{Our approach is SOTA against other methods and a MUSIQ-multi baseline.  Combining multi-scale training and OFA gives better performance even at lower resolutions.}}
\label{tab:sota}
\end{table}

\begin{table}[t]
\centering
\begin{tabular}{l|l}
Method    & mAP  \\ \hline
VGG-16 \cite{simonyan2015very}   & 79.3 \\
Swin-B  \cite{liu2021swin}  & 84.9 \\
Deit-B \cite{wu2021rethinking}   &  83.0 \\
ViT-B \cite{ViT}    & 81.7 \\
PF-DLDL \cite{gao2017deep} & 92.4	 \\
MCAR \cite{gao2021learning} & 94.3 \\
\highlightx{MUSIQ-multi} & 90.2 \\
MUSIQ-multi  + MAE + OFA & \textbf{95.4}
\end{tabular}
\caption{Results of multilabel classification over 20 classes on Pascal VOC2012.}
\label{table:Pascal}
\end{table}

\begin{figure}[t]
\centering
\includegraphics[width=0.6\textwidth]{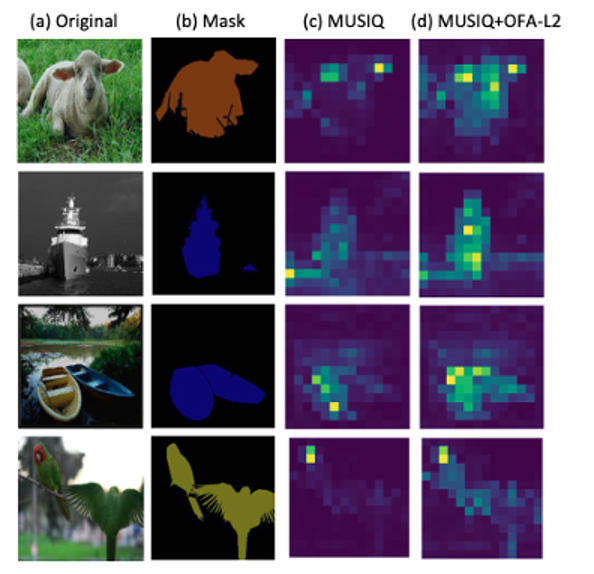}
\caption{Comparison of attention maps of proposed MUSIQ + OFA and baseline MUSIQ.}
\label{fig:attmaps1}
\end{figure}

In Figure \ref{fig:attmaps1}, we visualize the final-layer attention maps of the baseline MUSIQ and MUSIQ + OFA for some test examples.  We find that MUSIQ + OFA qualitatively attends to object shapes more consistently and produces reasonable segmentation maps in comparison to MUSIQ.  This finding is consistent across small-single label images, large single-label images, multi-label images, and multi-label multi-object images. MUSIQ often attends more greatly to the background and finds spurious correlations through attention while the OFA loss has a significant impact in focusing the attention computation on objects and greatly reducing attention to the background.
%although it too attends to background context in some cases.

\begin{figure}[t]
\centering
\includegraphics[width=0.6\textwidth]{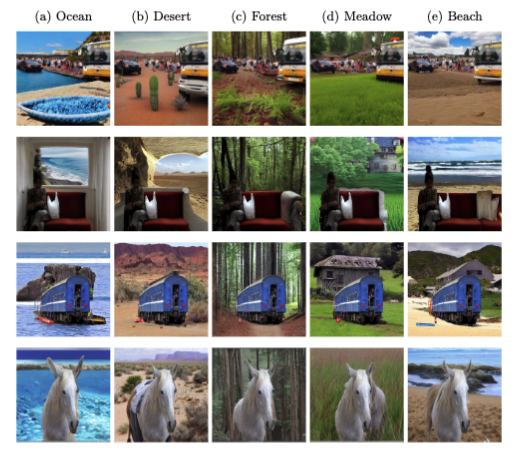}
\caption{Example images generated by Stable Diffusion inpainting on MS COCO.}
\label{fig:sdimg}
\end{figure}

Table~\ref{table:Pascal} compares the performance of zero-shot MUSIQ-multi  + MAE + OFA (trained on MS COCO), to recent SOTA transformers: ViT-B \cite{ViT}, Swin  (Swin-B) \cite{liu2021swin}, DeiT with iRPE-K (DeiT) \cite{wu2021rethinking}, PF-DLDL \cite{gao2017deep}, MCAR \cite{gao2021learning} and to VGG-16 \cite{simonyan2015very}.
Our model exhibits the best performance and significantly outperforms the other methods.

\begin{figure}[h]
  \centering
  \begin{minipage}{0.5\textwidth}
    \centering
    \includegraphics[width=1\textwidth]{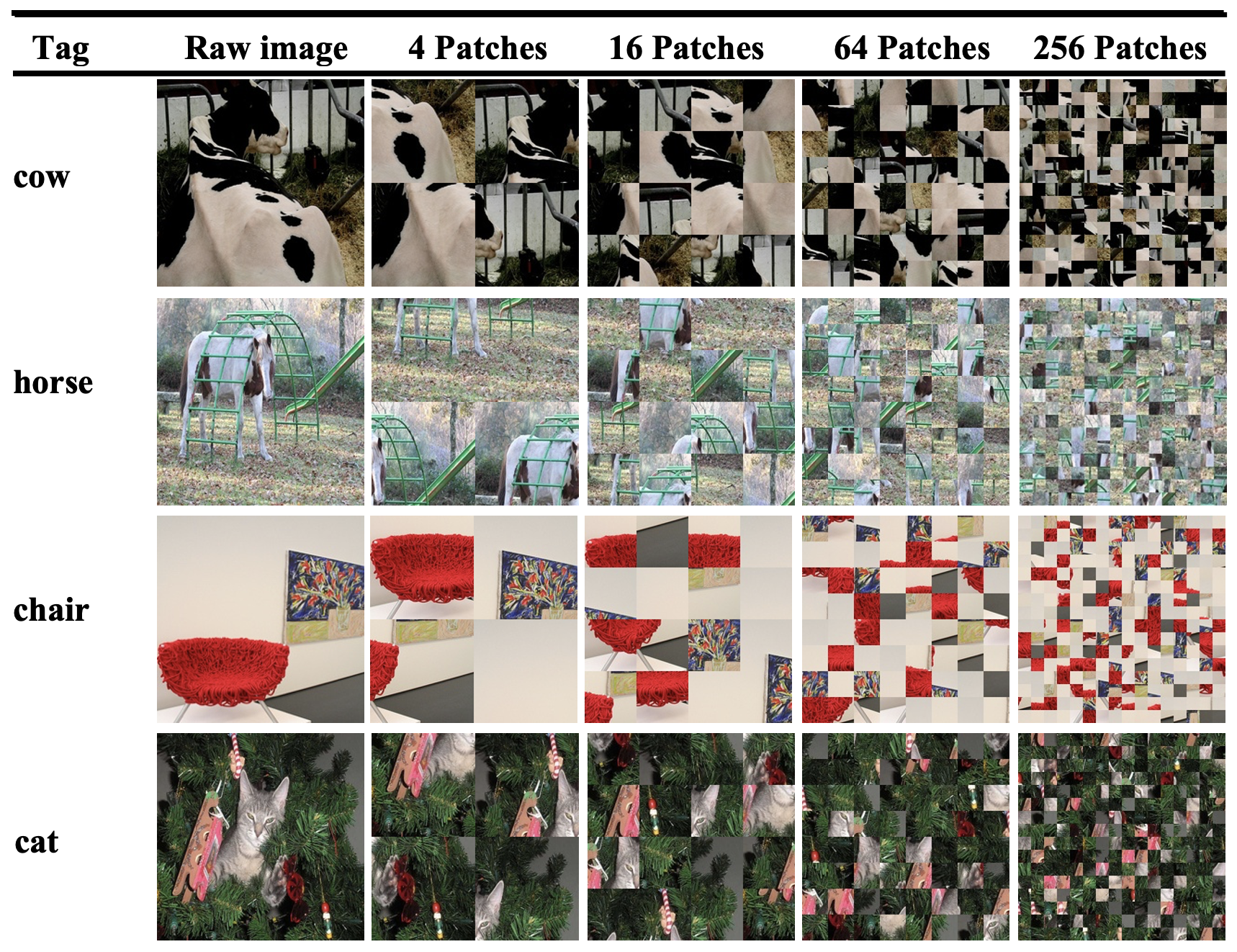}
    %\caption{Example shuffle operation applied to a varying number of patches. For humans the objects in a shuffled grid with 4 patches already seem unrecognizable.}
    %\label{fig:shuffle}
 \end{minipage}\hfill
  \begin{minipage}{0.48\textwidth}
    \centering
    \includegraphics[width=1\textwidth]{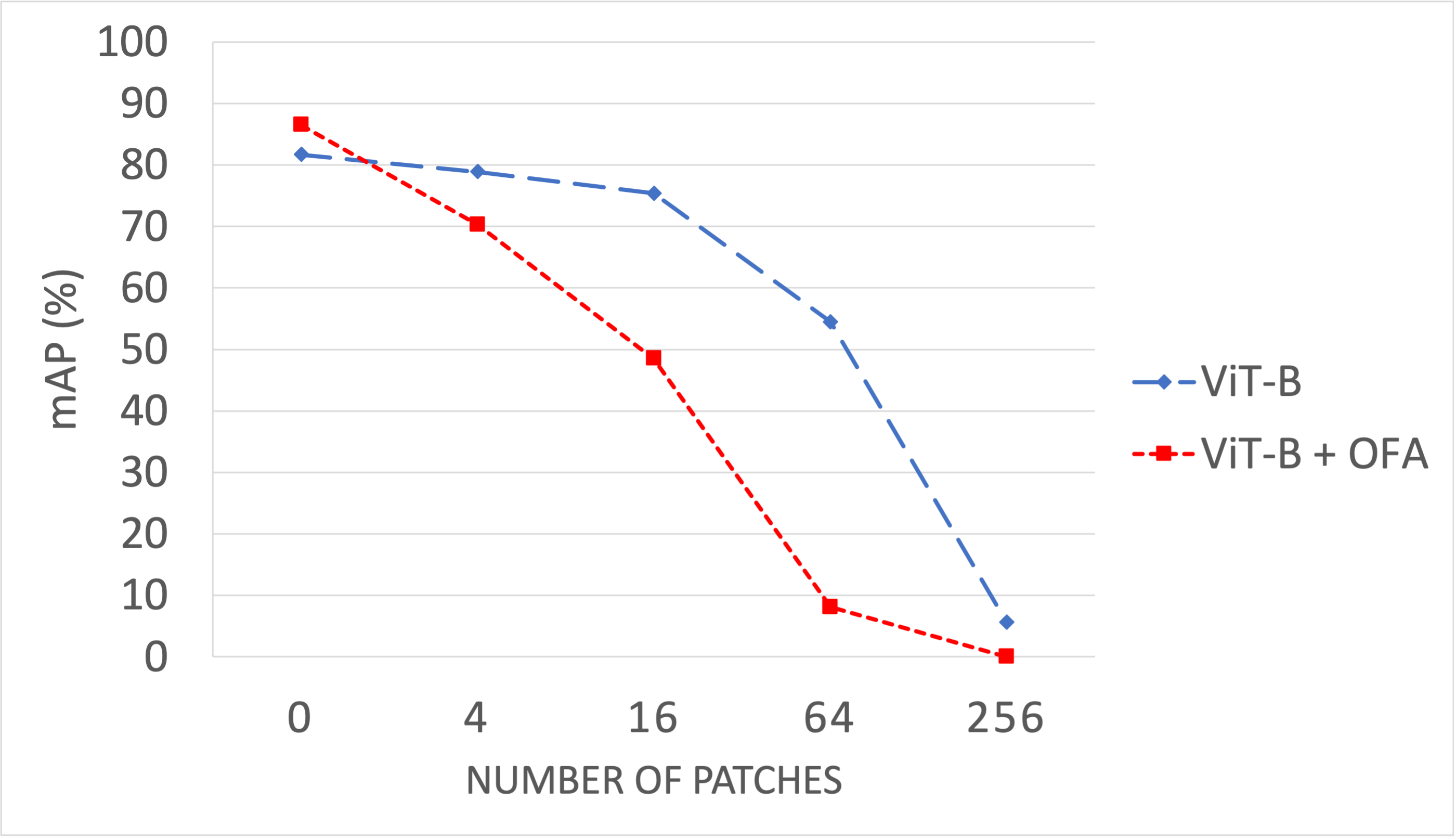}
    \caption{Example shuffle operation applied to a varying number of patches. For humans, the objects in a shuffled grid with 4 patches already seem unrecognizable.
    The mAP over 20 classes on PASCAL VOC2012 when patches are shuffled. While the classification performance of ViT + OFA drops significantly, those of ViT hardly drops.}
    \label{fig:shuffleresults}
  \end{minipage}
\end{figure}

\begin{table*}
\centering
\begin{tabular}{c|c|c|c|c}
Base Model              & Resolution & Baseline ViT & ViT+OFA   \\ \hline
ViT-Base-Patch16 (1k)   & 224        & 73.9 (-7.0)  & \textbf{78.6 (-2.2)} \\
ViT-Base-Patch16 (21k)  & 224        & 73.6 (-9.3)  & \textbf{81.7 (-2.2)} \\
ViT-Large-Patch16 (21k) & 384        & 79.0 (-6.9)  & \textbf{83.7 (-3.0)}
\end{tabular}
\caption{mAP results on MS COCO test data with background in-painted by Stable Diffusion \cite{StableDiffCVPR2022}. We show the performance on the original test set and the degradation on our inpainted dataset. The OFA model is more robust to background perturbations.  The result implies that OFA is more focused on learning semantic information about the objects rather than spurious correlations to the background.}
\label{tab:stablediffusion}
\end{table*}

\footnotesize
\begin{table*}[h]
\centering
\footnotesize
\begin{tabular}{c|ccccccccccccc|c}
OFA at Different Layers (40\% data)     & 1 & 2 & 3 & 4 & 5 & 6 & 7 & 8 & 9 & 10 & 11 & 12 &                                & mAP           \\ \hline
{[}12{]}           &   &   &   &   &   &   &   &   &   &   &   & \checkmark &  & 83.5        \\
{[}1{]}            & \checkmark &   &   &   &   &   &   &   &   &   &   &   &  & 83          \\
{[}1,12{]}         & \checkmark &   &   &   &   &   &   &   &   &   &   & \checkmark &  & 83.6        \\
{[}1,6,12{]}       & \checkmark &   &   &   &   & \checkmark &   &   &   &   &   & \checkmark &  & 83.7        \\
{[}1,3,7,10,12{]}  & \checkmark &   &   & \checkmark &   &   & \checkmark &   &   & \checkmark &   & \checkmark &  & \textbf{84.0} \\
{[}1,3,5,7,9,11{]} & \checkmark &   & \checkmark &   & \checkmark &   & \checkmark &   & \checkmark &   & \checkmark &   &  & 83.7        \\
{[}all{]}          & \checkmark & \checkmark & \checkmark & \checkmark & \checkmark & \checkmark & \checkmark & \checkmark & \checkmark & \checkmark & \checkmark & \checkmark &  & 83.6        \\ \hline
\end{tabular}
\caption{\highlightx{Ablation of computing OFA loss on multiple attention blocks in ViT+OFA using the ViT-Base-Patch16 (21k) on a subset of MS COCO.}}
\label{tab:ofaablation}
\end{table*}

\subsection{Out-of-Distribution Background Corruption with Stable Diffusion}
\label{sec:ood}
In Figure \ref{fig:sdimg}, we show selected examples of our new dataset for evaluation of OFA on OOD samples with adversarially corrupted backgrounds. We use Stable Diffusion inpainting \cite{StableDiffCVPR2022} to replace backgrounds in each of the MS COCO test images with five new background categories: ocean, desert, forest, meadow, and beach. We use the mask information for each image to set boundaries for parts of the image that are inpainted.  We inpaint the background of each image while leaving the object area unaltered, effectively superimposing each object onto a new background.  To decide on the inpainting domain we use the simple prompts to guide the diffusion process.  We use 5 prompts for each validation image resulting in an overall set of $5 \times 5000 = 25,000$ images.
We then test models trained on MS COCO without any finetuning.  
Table \ref{tab:stablediffusion} clearly shows the robustness of OFA to OOD images with respect to background perturbations.  
We find that ViTs are susceptible to background perturbations showing a significant decrease in performance while the OFA model is more robust to background swapping.

\subsection{Learning Shape Representations over Textures}
We demonstrate that the arrangement of object parts is not well represented by a standard ViT and is aided by using OFA. We divide an input image into patches by imposing a grid structure of different sizes and then randomly permute the position of patches.
Fig.~\ref{fig:shuffle} shows samples of this shuffle operation applied to PASCAL VOC 2012 images \cite{pascal-voc-2012}.
As illustrated by the blue dashed curve in Fig.~\ref{fig:shuffleresults}, the multilabel classification performance of ViT remains nearly constant if 4 patches and 16 patches are permuted.
However, as can be seen in Fig.~\ref{fig:shuffle},
already the objects in the images with 4 permuted patches seem unrecognizable to a human.
If ViT possessed an understanding of the configural shape, 
we should see a significant performance drop.
In contrast, the performance of ViT + OFA drops significantly (red dashed curve).
This demonstrates that it gained at least a rough understanding of configural object shapes due to the object-focused attention loss.
We used ViT as the baseline model in this experiment 
to eliminate any influence of multi-scale and aspect ratio preserving since ViT takes a single-scale, square image of size $256 \times 256$ as input.

\section{Discussion and Future Work} 
\label{sec:discussion}
We introduce a simple yet effective method for object-centered learning in the vision transformer framework. The proposed object focus attention loss is easily integrated into the self-attention module.  Our trained model does not introduce any computational overhead at inference and still outperforms SOTA transformers.  Moreover, it generalizes better to out-of-distribution examples and corrupted examples with respect to background and object shape.  Finally, we show SOTA results when our approach is combined with multi-scale representation and MAE, offering a potential avenue for more exploration.  We are interested in scaling our method to larger data using models that generate pseudo-segmentation masks such as SAM. We will explore this option in our future work.  As shown in \cite{Kellman2018,GeirhosNeurIPS2018,texture-shape-ICLR2019},
deep learning models tend to focus on texture rather than on the shape of objects.  Our experimental results demonstrate that the proposed refocusing of attention on segmentation masks contributes to a better understanding of holistic object shapes.
We speculate that this fact makes our model more robust to adversarial attacks.  In order to refine the learned attention, we will also consider learning attention based on instance segmentation as well as on panoptic segmentation data.

% \newpage
\bibliographystyle{splncs04}
\bibliography{transformer}
%





% LNCS Homepage, \url{http://www.springer.com/lncs}. Last accessed 4
% \end{thebibliography}
\end{document}